\documentclass[runningheads]{llncs}
\usepackage{graphicx}
\usepackage{cite}
\usepackage{amsmath,amssymb,amsfonts}
\usepackage{algorithmic}
\usepackage{textcomp}
\usepackage{xcolor,comment,subcaption}
\usepackage{arabtex,utf8}
\usepackage{todonotes}
\begin{document}
\title{The Who in Code-Switching: A Case Study for Predicting Egyptian Arabic-English Code-Switching Levels based on Character Profiles}
\titlerunning{Predicting Code-Switching Levels based on Character Profiles}
%
\author{Injy Hamed \inst{1} \and Alia El Bolock\inst{2,3} \and
Cornelia Herbert\inst{2} \and Slim Abdennadher \inst{3} \and \\Ngoc Thang Vu \inst{1}}
\authorrunning{Hamed et al.}
%

\institute{Institute for Natural Language Processing, University of Stuttgart, Stuttgart, Germany \and
Applied Emotion and Motivation Psychology, Ulm University, Ulm, Germany \and  Informatics and Computer Science, German International University, Cairo, Egypt
\email{\{injy.hamed,ngoc-thang.vu\}@ims.uni-stuttgart.de} \\
\email{\{alia.elbolock,slim.abdennadher\}@giu-uni.de}\\
\email{cornelia.herbert@uni-ulm.de}
}

\maketitle              
\begin{abstract}
Code-switching (CS) is a common linguistic phenomenon exhibited by multilingual individuals, where they tend to alternate between languages within one single conversation. CS is a complex phenomenon that not only encompasses linguistic challenges, but also contains a great deal of complexity in terms of its dynamic behaviour across speakers. Given that the factors giving rise to CS vary from one country to the other, as well as from one person to the other, CS is found to be a speaker-dependant behaviour, where the frequency by which the foreign language is embedded differs across speakers. While several researchers have looked into predicting CS behaviour from a linguistic point of view, research is still lacking in the task of predicting user CS behaviour from sociological and psychological perspectives. We provide an empirical user study, where we investigate the correlations between users' CS levels and character traits. We conduct interviews with bilinguals and gather information on their profiles, including their demographics, personality traits, and traveling experiences. We then use machine learning (ML) to predict users' CS levels based on their profiles, where we identify the main influential factors in the modeling process. We experiment with both classification as well as regression tasks. Our results show that the CS behaviour is affected by the relation between speakers, travel experiences as well as Neuroticism and Extraversion personality traits.
\end{abstract}
\section{Introduction}
Code-switching (CS), the act of alternating between two or more languages within a conversation, has become a common linguistic phenomenon among speakers in bilingual/multilingual societies. CS has become a worldwide multilingual phenomenon, which has evolved due to several different factors. In the middle east, colonization as well as international businesses and education have played a major role in introducing English and French into everyday conversations. CS has become dominant in many countries including Arabic-French in Morocco \cite{Ben83} and Algeria \cite{BD83}, Arabic-English in Egypt \cite{Abu91}, Saudi Arabia \cite{OI18}, Jordan \cite{MA94}, Kuwait \cite{Akb07}, Oman \cite{AlQ16}, UAE \cite{Khu03}, and both Arabic-English and Arabic-French in Lebanon \cite{BB11} and Tunisia \cite{Bao09}. Code-switching is usually defined as a mixture of two distinct languages: primary/matrix language and secondary/embedded language. 
CS can occur at the levels of sentences, words, as well as morphemes in the case of highly morphological languages, such as Arabic, presenting the following CS types:
\setcode{utf8}
\begin{itemize}
    \item Inter-sentential CS: where switching occurs at the sentence-level. For example:\\
    ``It was really nice. \<اتعلمت كتير> .''
    \newline(\textbf{It was really nice.} I learnt a lot.)
    \item Intra-sentential CS: where switching occurs at the word-level. For example:\\
    ``I do not think \<أني عاوز أبقى> student anymore.''  \newline(\textbf{I do not think} I want to be a \textbf{student anymore}.)
    \item Extra-sentential CS: where a loan word is borrowed from the secondary language. For example:\\
    ``\<حلوة جدا.> experience \<كانت>''\newline (It was a very nice \textbf{experience}).
    \item Intra-word CS: where switching occurs at the level of morphemes. For example:\\
    ``.[conference \<ال>] 
    [target \<هن>]
    \<احنا>''
    \newline(We [will \textbf{target}] [the \textbf{conference}])

\end{itemize}
While research in this field goes back to 1980, where Poplack \cite{Pop80} presented influential theories on CS, there is still room for many research directions to be explored. CS has gained most attention with regards to its linguistic perspective, where it has been extensively studied \cite{Mye97,SP81,KZ18} and several approaches were proposed to integrate linguistic findings into NLP systems \cite{LF12,LF14a,LF14b,AKT+14}. It has also received considerable attention from socio- and psycho- linguists \cite{HA01,Che03,RB13,Eld14,DW14,DL14,Ben17,EKA+20}, identifying factors affecting this behaviour. We have also seen an increasing interest in the field of computational linguistics to build NLP systems that can handle such mixed input \cite{CV16,hamed2017building,TP18,AAS+18,YLY+19,LV19,HZE+19,SCR+19,SIA20,CSE+20,hamed2022investigations,hamed2022investigating,ACH+21,CHA+21,HCA+22}. 
Only few researchers have worked on building user-adaptive NLP systems that take into consideration users' CS styles. It was shown, however, that incorporating users' different CS attitudes achieves significant improvements in NLP tasks, such as language modeling and speech recognition \cite{VAS13,RSB18}. We are thus motivated to further analyze the CS behaviour from socio- and psycho-linguistic perspectives and examine the effectiveness of using users' profiles to predict their CS levels. This information can then be used as a first-pass to user-adapted NLP systems.\\

In this paper, we advance the current state-of-the-art by investigating the correlation between users' CS behaviour and their character profiles. As proposed by Character Computing \cite{elcharacter}, behavior is driven by character, i.e. traits (e.g., personality, socio-demographics, and background) and states (e.g., affect and health) and situation \cite{el2020character}. Character, i.e., a person's profile interacts with the individual's current situation and influences their behavior ~\cite{herbert2020experimental}. We hypothesize that the concepts of character computing can be used to infer speakers' CS behavior based on their character profiles. In the scope of our work, we look into the following research questions: a) does character affect one's CS behaviour? If yes, which character traits are the most influential? and b) with prior user knowledge, can we predict the CS level?\\

In order to answer these questions, we conduct an empirical study where we collect a novel dataset containing participants' CS levels as well as character traits.
We rely on participants' CS actual levels rather than self-reported figures, where we hold interviews with the participants, and calculate CS levels from interviews' transcriptions. We then identify correlations and validate them using machine learning, by looking into the factors identified in the modeling process to have high predictor importance. We present the overall plan of our work in Figure \ref{fig:outline}. \\

In this work, we make the following contributions: (1) provide a corpus containing participants' CS levels and character meta-data, thus paving the way for further analyses into CS in general and Arabic-English CS in specific; (2) provide further insight into the correlation between CS and personality traits, which hasn't been thoroughly investigated; (3) bridge the gap between the theories presented by socio- and psycho-linguists and computational linguistics research in the field of code-switching; and (4) present the first efforts in using machine learning to model users' CS behavior based on character profiles. We extend our work in \cite{hamed2021predicting} with further experiments and analyses. Our experiments show promising results with regards to predicting users' CS styles given certain character traits. We show that CS is affected by the roles and relationship of participants. We also identify the most influential factors in users' profiles to be traveling experiences, neuroticism and extraversion. 
\section{Related Work}
CS has presented itself as an interesting research area in different fields, including linguistics, socio- and pschyo- linguistics, as well as computational linguistics. Researchers have worked on understanding the CS behaviour and have shown that it is not a stochastic behaviour, but is rather governed by linguistic constraints, and affected by sociological and psychological factors. Such findings provide further understanding of the phenomenon and allow computational linguists to improve Natural Language Processing (NLP) applications accordingly. In this section, we present previous work covering factors affecting users' CS behaviour, as well as work that has integrated linguistic factors and users' CS attitudes/styles into NLP systems.

\subsection{Factors Affecting CS}
As stated in \cite{VAS13}, CS is a speaker-dependent behaviour, where the ``CS attitude'' differs from one person to the other. Researchers have identified several factors affecting users' CS behaviour, where it has been attributed to various linguistic,  psychological, sociological, and external factors.\\

From a linguistic perspective, several studies have looked into where switch points occur. Researchers identified CS POS tags and trigger words which most likely precede a CS point \cite{SFB09,BGS+18,HEA18,balabel2020cairo,hamed2020arzen}. In \cite{HEA18,balabel2020cairo,hamed2020arzen}, CS POS tags and trigger words were investigated for Egyptian Arabic. It was found that CS occurs most commonly for nouns, followed by verbs. It was also found that CS most commonly occurs after the Arabic definite article as well as after conjugations. The case of CS occurring after definite articles is common in the case of extra-sentential code-switching. It is a common case of morphological code-switching, where bilinguals attach the Arabic definite article \textit{Al-} to noun loan words. For the case of CS occurring after conjugations, it is reasonable since conjugations join two parts of sentences. Researchers also examined grammatical constraints providing information on where code-switching points are allowed in a sentence \cite{SP81,BRT94}. The Equivalence Constraint \cite{SP81} and Matrix constraint \cite{Mye97} are among the most popular linguistic theories. Researchers also used machine learning to predict code-switching points \cite{SL08}. \\

The sociological and psychological factors affecting CS behaviour have been less examined by computational linguists, however, they have been extensively studied covered by socio- and psycho-linguistics. Researchers have attributed CS behaviour to several factors:
\begin{itemize}
    \item Language preference: Bilinguals are driven towards code-switching whenever the second language is linguistically easier, for example, whenever a word is not accessible in the first language \cite{HA01,TBA+13,Che03}, and whenever some words are easier, more distinguishable and easier to use or the concepts involved are easier to express in that languages \cite{Che03}.
    \item External factors: CS attitude is affected by external factors, where it is affected by the Participant Roles and Relationship (whom the person is addressing) \cite{RB13}. It is also affected by the topic of conversation \cite{Vel10,RB13,EKA+20}. In \cite{Vel10}, it was found that code-switching stood out in certain topics, including family, school, ethnicity, and friends. In \cite{EKA+20}, it was found that CS percentage of CS varies across the following topics (in descending order): interests, describing pictures, personality, background, travel.
    \item Social factors: CS is also affected by social factors, such as age, gender, religion, level of education and social class \cite{Ben17,Rih,RB13}. In our study, occupation is found to greatly affect CS behaviour, where teaching assistants alternate between languages nearly twice as much as students. Such an observation could be justified by the nature of their jobs. These findings are important to further understand and model CS behaviour.
    \item Personality traits: Less work has investigated the link between CS and personality traits. In~\cite{DW14}, the link between Extraversion and Neuroticism with CS attitude is investigated, and Neuroticism is found to have a strong effect, while Extraversion was reported to have no effect. In~\cite{DL14}, Extraversion is found to be significantly linked to CS attitude. It is to be noted that these works were based on self-reported CS levels. 
    \item Emotions: It was found that individuals may code-switch to express certain feelings and attitudes \cite{Eld14} or to distant themselves from emotional events \cite{HA01}.
    \item Done deliberately for certain purposes: It was also found that code-switching can be done intentionally for the speaker's own benefit. CS can be used to capture attention \cite{Che03,Hol17}, reflect a certain socioeconomic identity which can give the speaker more credibility and reliability \cite{Ner11}, persuade an audience \cite{Hol17}, appeal to the literate/illiterate \cite{Che03}. It has also been agreed by several researchers that a speaker may code-switch intentionally to express group solidarity \cite{Che03,TBA+13,Eld14,Rih} or reflect social status \cite{Eld14}. As stated by Peter Auer \cite{Aue13}, ``Code-switching carries a hidden prestige which is made explicit by attitudes''. It is also reported that CS can be used for excluding another person from the dialog \cite{Che03}. While these factors may be attributed to some psychological factors, the connection between CS levels and personality traits is still unclear. 
\end{itemize}
\subsection{Adaptive NLP Systems}
Computational linguists have integrated findings from CS linguistic studies into NLP models. In \cite{AKT+14}, different factors were explored for applying factored language models for CS speech recognition task, where the factors were chosen such that they can provide strong predictions of code-switching points, such as POS and LID. Researchers have also investigated incorporating linguistic constraints for CS into language models to pose constraints on code-switch boundaries. In \cite{LF12,LF14a,LF14b},  the ``Equivalence Constraint'' introduced by Sankoff and Poplack \cite{SP81} as well as the ``Functional Head constraints'' developed by Belazi et al. \cite{LF14b} were integrated into language modeling. In \cite{RSG+21}, the authors developed the GCM toolkit for generating synthetic CS data based on the equivalence constraint. Researchers have used the GCM toolkit for data augmentation for the tasks of machine translation \cite{JNA+21} as well as language modeling and speech recognition \cite{HCA+22}.\\

Few work has been done towards building adapting NLP systems towards different CS attitudes and styles of bilingual speakers. In \cite{VAS13}, it was shown that clustering speakers according to their code-switching attitude can lead to language models that model code-switching more precisely. By analyzing the CS Mandarin-English SEAME corpus, it was found that whether a POS tag acts as a trigger for switching languages or not depends on the speaker. Accordingly, the training data was clustered into different classes, representing different code-switching attitudes. The training data clusters were then used to adapt the language model. The adapted models showed significant reduction in perplexity, as well as improvement in MER. 

In \cite{RSB18}, features extracted from acoustics were used to distinguish between different code-switching styles, where style-specific language models showed reduction in perplexity. Previous research \cite{NL00} has also demonstrated the importance of taking personality into account when building speech user interfaces systems. It was shown that users preferred computer-generated speech that exhibits similar personality to their own, indicating such speech as more attractive, credible and informative. 

Despite previous work demonstrating the importance of adapting NLP systems as well as the various factors highlighted by socio- and psycho- linguists to be affecting the CS phenomenon, up to our knowledge, this is a current gap in research which has not been investigated in CS NLP systems. 
A pre-condition to building user-adaptive systems is understanding the code switching behavior of users. In this paper, we examine the effect of different character traits on users' CS behaviour as well as the effectiveness of using users' profiles to predict their CS usage using machine learning.

\section{Experimental Setup}
\begin{figure}[t]
  \caption{The outline of the work.}
  \label{fig:outline}
  \centering
  \includegraphics[width=\linewidth]{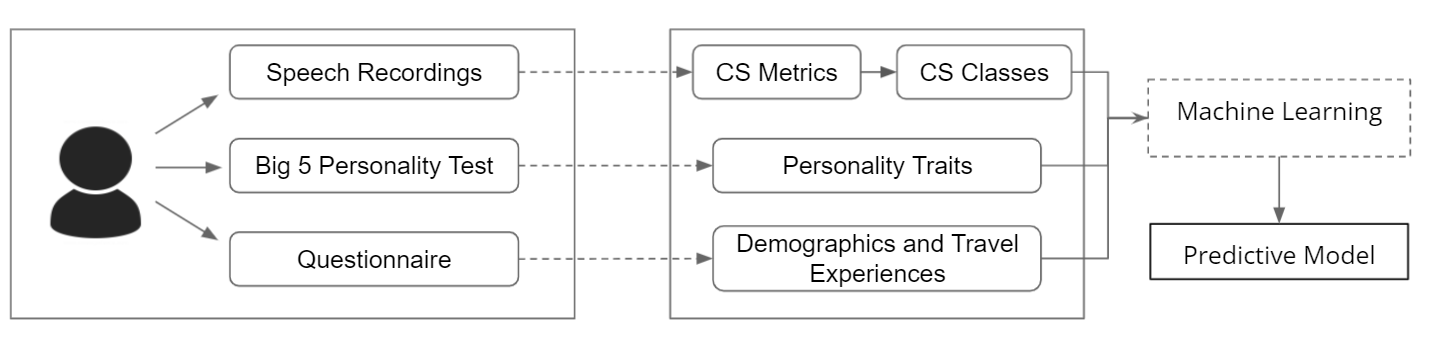}
\end{figure}

In this section, we outline the experimental settings. In Figure \ref{fig:outline}, we present the overall plan of our work. The work involves three main phases: (1) Collecting users’ information including CS usage, demographics, personality traits and traveling experiences; (2) Categorizing users into classes and dividing the dataset into balanced train and test sets; and (3) Building predictive models using machine learning.

\subsection{Data Collection}
Our first contribution is to collect a corpus containing quantified measures of participants'  CS levels and participants' meta-data. In our study, we believe it is very important to rely on participants' actual levels of CS usage rather than self-reported levels. This was done by holding interviews where the following data is gathered: (1) demographics through a questionnaire, (2)  personality traits through the Big Five Personality test \cite{Gol92}, and (3) CS frequency from interview transcriptions. This corpus is made public in order to motivate further research in this direction\footnote{The data can be obtained by contacting the authors.}.\\

The participants included 65 Egyptian Arabic-English bilingual speakers. All participants are fluent Arabic-English bilinguals. The interview covered general topics, including technology, education, hobbies, traveling, and life experiences. In order to avoid external factors affecting the participants' CS behaviour, each interview involved one interviewee and two interviewers (a male and a female). No instructions were given to participants regarding code-switching; they were not asked to produce nor avoid code-switching. For each participant, we collect socio- and psycho-linguistic meta-data, where participants were asked to fill a questionnaire and the Big Five Personality Test. We elaborate more on both forms in the following sections.

\subsubsection{Questionnaire}
The questionnaire gathers information about users including demographics (age, gender, occupation), history (educational background and traveling experiences), CS background (participants' levels of CS self-awareness and CS levels among families and friends) as well as users' CS perceptions. The questionnaire is provided in \ref{sec:appendix-questionnaire}.

\paragraph{Demographics}
The participants were gender balanced; 54\% were males and 46\% were females. All Participants were in the age range of 18-35, where 52.3\% were students and 47.7\% were university employees. On a survey on the language the participants considered as their mother tongue, 61.5\% of the participants identified their mother tongue as “Pure Arabic”, 0\% as “Pure English”, and 38.5\% as “Code-switched, Arabic-English”. We also asked the participants to provide the longest duration in which they resided in a foreign country for tourism/education. Participants' responses covered all provided duration intervals; ``I did not travel'' (18.5\%), ``Less than a month''(29.2\%), ``1-3 months''(15.4\%), ``3-6 months''(23.0\%), ``6-12 months''(6.2\%), ``1-3 years''(6.2\%), ``more than 3 years''(1.5\%). For travelling experiences, we did not consider the languages of the countries the participants traveled to, but rather the stay duration. It is however common that people use English as a common language across different countries. We also gathered information about the participants' education. We report that 86\% of the participants reported that they had classes at school where speaking English was obligatory. Also, 61.5\% of the participants attended national schools, while 38.5\% attended international schools. In national schools, the governmental curriculum is taught either in Arabic/foreign language depending on whether the school is a regular/language school. In international schools, an international curriculum is taught in a foreign language. Given that it is common for people to confuse the terms ``international schools'' with national ``language schools'', we did not include this feature in our analysis. 

\paragraph{Participants' CS usage} On a scale of 1-5, participants were asked to rate their frequency of CS as well as how aware they are of their CS usage. Participants report an average rating of 3.7 for CS frequency (with a minimum rating of 2, reflecting that they all code-switch) and usage awareness rating of 4.0. We also gathered information about the language use of family and friends. For 83\% of the participants, their family members speak more than one language. This is expected as the family members of the participants involved in this experiment received their education in the post-colonial era, in which English was a mandatory subject in schools. The percentage is higher among friends, reaching 94\% of the participants.\\

While in the scope of this paper, we will focus on the correlation between CS and certain factors (personality, age, gender, occupation, and travel experiences), our corpus contains more information and provides a valuable resource where the CS phenomenon can be further investigated. 

\subsubsection{Personality Test}
The Big Five Personality Test: The test was chosen as it is the most widely used and extensively researched model of personality \cite{GRS03} and because it consists of 50 questions, requiring around 10 minutes. The Big Five Personality Test assesses five major dimensions of personality:
\begin{itemize}
    \item Openness: This trait features characteristics such as imagination and insight. People who are high in this trait also tend to have a broad range of interests. They are curious about the world and other people and eager to learn new things and enjoy new experiences.
    \item Conscientiousness: Standard features of this dimension include high levels of thoughtfulness, good impulse control, and goal-directed behaviors. Highly conscientious people tend to be organized and mindful of details. They plan ahead, think about how their behavior affects others, and are mindful of deadlines.
    \item Extraversion: Extraversion (or extroversion) is characterized by excitability, sociability, talkativeness, assertiveness, and high amounts of emotional expressiveness. People who are high in extraversion are outgoing and tend to gain energy in social situations. Being around other people helps them feel energized and excited.
    \item Agreeableness: This personality dimension includes attributes such as trust, altruism, kindness, affection, and other prosocial behaviors. People who are high in agreeableness tend to be more cooperative while those low in this trait tend to be more competitive and sometimes even manipulative.
    \item Neuroticism: Neuroticism is a trait characterized by sadness, moodiness, and emotional instability. Individuals who are high in this trait tend to experience mood swings, anxiety, irritability, and sadness. Those low in this trait tend to be more stable and emotionally resilient.
\end{itemize}
In Table \ref{table:personalityTraitsValues}, we show the mean and standard deviation of participants' scores across the five personality traits. For the experiments, all personality trait scores are normalized using the z-score.
\begin{table}[t]
  \caption{Statistics on participants' scores on the personality traits.}
  \label{table:personalityTraitsValues}
  \centering
  \setlength\tabcolsep{2pt}
  \resizebox{\columnwidth}{!}{
  \begin{tabular}{|l|r|r|r|r|r|}
    \hline
    \multicolumn{1}{|c|}{\textbf{}} & \multicolumn{1}{c|}{\textbf{Openness}} & \multicolumn{1}{c|}{\textbf{Conscientiousness}} & \multicolumn{1}{c|}{\textbf{Extraversion}} & \multicolumn{1}{c|}{\textbf{Agreeableness}} & \multicolumn{1}{c|}{\textbf{Neuroticism}} \\\hline
    \textbf{Mean} & 35.5 & 31.8 & 24.5 & 34.0 & 24.6 \\\hline
    \textbf{STDEV} & 6.6 & 5.0 & 4.2 & 4.9 & 6.3 \\\hline

    \end{tabular}
    }
\end{table}

\subsubsection{Users' CS Frequency}
The interviews were manually transcribed by professional transcribers. The transcriptions contain 8,339 utterances, having 143,480 tokens. We use three metrics for measuring participants' CS usage: (1) percentage of monolingual English sentences (reflecting the level of inter-sentential CS, where CS occurs at sentence-level), (2) percentage of CS sentences, and (3) Code-Mixing Index (CMI) \cite{DG14} (reflecting the level of intra-sentential CS, where CS occurs at word-level). CMI is defined as:
\[CMI=\frac{\sum_{i=1}^{N}(w_i)-max\{w_i\}}{n-u} \]
where $\sum_{i=1}^{N}(w_i)$ is the total number of words over all languages, $max(w_i)$ is the highest number of words across the languages, $n$ is the total number of words, and $u$ is the total number of language-independent words. Monolingual sentences would have a CMI of 0 and sentences with equal word distributions across languages would have a CMI of $n/N$, which is 0.5 in the case of bilingual utterances. In Table \ref{table:CS_stats}, we report participants' CS statistics with regards to the three metrics. 
\begin{table}[t]
  \caption{Statistics on participants' CS usage.}
  \label{table:CS_stats}
  \centering
  \setlength\tabcolsep{2pt}
  \resizebox{0.7\columnwidth}{!}{
  \begin{tabular}{|l|r|r|r|r|}
    \hline
    \multicolumn{1}{|c|}{\textbf{CS Metrics}} & \multicolumn{1}{c|}{\textbf{Min}} & \multicolumn{1}{c|}{\textbf{Max}} & \multicolumn{1}{c|}{\textbf{Mean}} & \multicolumn{1}{c|}{\textbf{STDEV}} \\\hline
    \textbf{CMI} & 0.06 & 0.47 & 0.17 & 0.10\\\hline
    \textbf{CS Sentences (\%)} & 31.25 & 85.45 & 62.16 & 12.71 \\\hline
    \textbf{English Sentences (\%)} & 0.00 & 13.33 & 3.76 & 3.76\\\hline
    \end{tabular}
    }
\end{table}

\subsection{Machine Learning Models}
We use \textit{IBM SPSS Modeler} tool \cite{MAB+13} to build our predictive models. We experiment on both tasks; classification and regression. For classification, we classify participants' CS levels into 4 classes, and investigate the ability of the ML models to classify participants into their correct classes. We use the following ML algorithms: Random Trees, Random Forest, XGBoost Tree, XGBoost Linear, and Linear Support Vector Machine (LSVM). We opt for the classification task, as building user-adaptive NLP systems would involve tailoring the systems for classes of user group, as in \cite{VAS13}, thus representing and predicting users on a CS class-level would be useful for this task. Also, in classification, multiple CS metrics can be used to define users' CS levels. We also present results on the regression task for further analyses. For the regression task, we experiment with predicting both percentage of English words as well as CMI using the following models: Random Trees, Random Forest, Linear Support Vector Machine (LSVM), Linear Regression, and Generalized Linear Regression.

\subsubsection{Categorizing Users' CS Levels}
\setcode{utf8}
\begin{table}[t]
    \caption{Examples of utterances from classes. The arrows denote the sentence starting direction, as Arabic is written right to left.}
    \label{table:utterance_examples}
    \centering
    \setlength\tabcolsep{1.5pt}
    \begin{tabular}{|c|r|r|}
        \hline
        \textbf{Class} & \multicolumn{1}{c|} {\textbf{Utterance}} & \textbf{CMI}\\\hline
        \textbf{0}& 
        \<كشغل يعنى برة أحسن يعنى>
        opportunities
        \< بس يعنى على الأقل ال>
        \textleftarrow
         & 0.09\\\hline
        \textbf{1}&
        \<فيه او كده>
        courses
        \<و كده بفكر اخد >
        web development 
        \<يعنى موضوع ال>
        \textleftarrow
        &0.23\\\hline
        \textbf{2}&
        \multicolumn{1}{l|}
        {\textrightarrow challenging enough to motivate me
        \<ان انا اشتغل و اتعلم حاجة يعنى>}
        &0.42\\\hline
        \textbf{3}&
        \multicolumn{1}{l|}
        {\textrightarrow I like the idea
        \<هى بت>
        target
        \<حاجات حوالينا فى ال >
        Society}
        &0.50\\\hline
    \end{tabular}
\end{table}
In this section, we discuss how we categorize users' CS levels into classes. The CS levels were categorized into 4 classes using several categorization approaches. In the first approach, we quantize the CMI range (0-1) into 4 classes. By trying different divisions, we chose the following CMI ranges (0-0.1,0.1-0.2,0.2-0.3,0.3-0.5). As shown in Table \ref{table:utterance_examples}, the ranges reflect different CS levels, where CS in class 0 is mostly present as extra-sentential CS (borrowing). In class 1, CS includes extra-sentential and slight intra-sentential. Class 2 includes extensive intra-sentential CS. Class 3 represents extensive intra-sentential CS, in addition to having many sentences where the primary language is English with embedded Arabic words, such as 
``I need it for communication with others \<يعنى>''.\\

\begin{figure*}[t]%
\caption{The participants' distribution across classes for each categorization approach.}
\label{fig:categorization}
\centering
   \resizebox{\linewidth}{!}{
\begin{subfigure}{.3\linewidth}
\includegraphics[width=\linewidth]{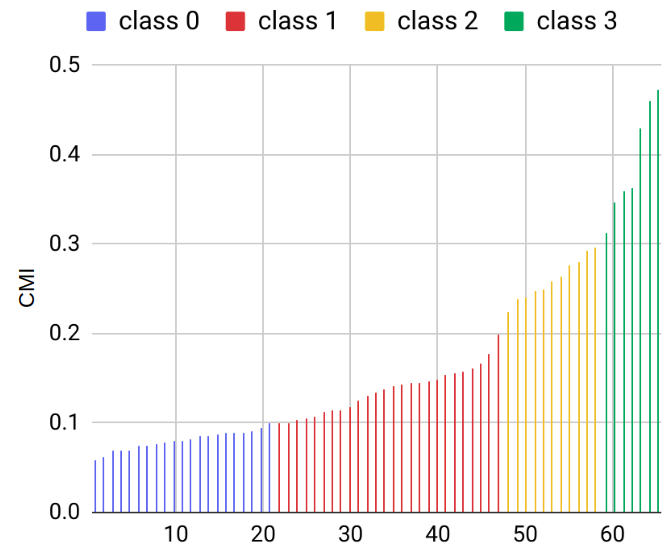}%
\caption{CMI}
\label{subfiga}
\end{subfigure}\hfill
\begin{subfigure}{.3\linewidth}
\includegraphics[width=\linewidth]{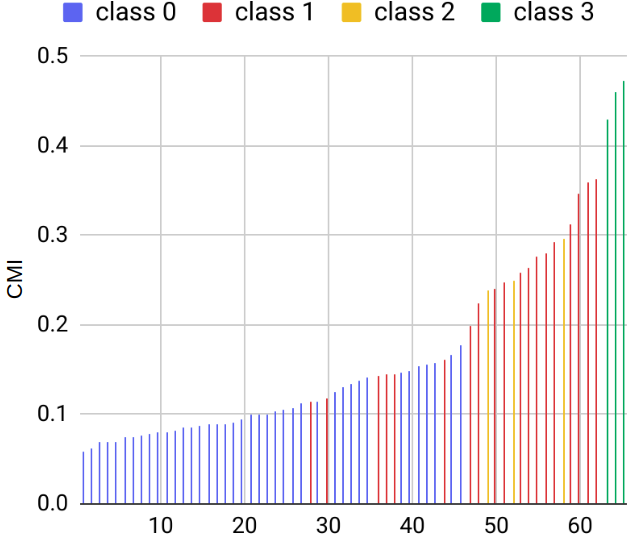}%
\caption{K-means[CMI]}
\label{subfigb}
\end{subfigure}\hfill
\begin{subfigure}{.35\linewidth}
\includegraphics[width=\linewidth]{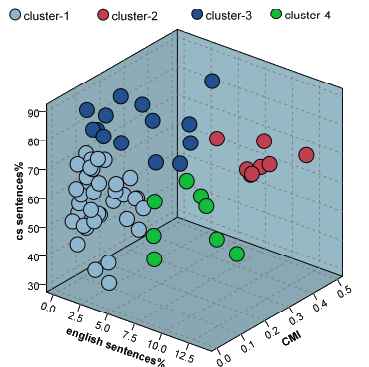}%
\caption{K-means[CMI+CS\%+En\%]}
\label{subfigc}
\end{subfigure}
}
\end{figure*}

In the second approach (K-means[CMI]), instead of relying on our own judgement for classifying CS levels, we use K-means clustering algorithm to cluster users into 4 classes relying on only the CMI values as input. In the third approach (K-means[CMI+CS\%+En\%]), we rely on all three CS metrics, where the categorization is performed using K-means algorithm based on CMI, percentage of CS sentences, and percentage of monolingual English sentences. The distribution of participants across classes are further shown in Table \ref{table:categorization_stats}. Figure \ref{fig:categorization} shows the participants' distribution across classes for each categorization approach.
\begin{table}[t]
  \caption{Statistics on participants' categorization into classes: average CMI and size of each class.}
  \label{table:categorization_stats}
  \centering
  \setlength\tabcolsep{1.5pt}
\resizebox{\columnwidth}{!}{
\begin{tabular}{|l|r|r|r|r|r|r|r|r|}
\cline{2-9}
 \multicolumn{1}{c|}{}& \multicolumn{2}{|c|}{\textbf{Class 0}} & \multicolumn{2}{c|}{\textbf{Class 1}} & \multicolumn{2}{c|}{\textbf{Class 2}} & \multicolumn{2}{c|}{\textbf{Class 3}} \\\hline
\multicolumn{1}{|c|}{\textbf{Categorization}} & \multicolumn{1}{c|}{\textbf{Mean}} & \multicolumn{1}{c|}{\textbf{Size}} & \multicolumn{1}{c|}{\textbf{Mean}} & \multicolumn{1}{c|}{\textbf{Size}} & \multicolumn{1}{c|}{\textbf{Mean}} & \multicolumn{1}{c|}{\textbf{Mean}} & \multicolumn{1}{c|}{\textbf{Mean}} & \multicolumn{1}{c|}{\textbf{Mean}} \\\hline
\textbf{CMI} & 0.08 & 32.3\% & 0.14 & 40.0\% & 0.26 & 16.9\% & 0.39 & 10.8\% \\\hline
\textbf{K-means[CMI]} & 0.11 & 70.8\% & 0.26 & 18.5\% & 0.35 & 6.1\% & 0.45 & 4.6\% \\\hline
\textbf{K-means[CMI+CS\%+En\%]} & 0.11 & 61.5\% & 0.21 & 23.1\% & 0.31 & 10.8\% & 0.45 & 4.6\%\\\hline
\end{tabular}
}
\end{table}

\subsubsection{Splitting Train and Test Sets}
We divide the dataset into train (70\%) and test (30\%) sets. The split has been done taking into consideration having balanced divisions, such that for all categorization approaches, the distribution of samples in each class is kept nearly the same as the overall distribution, as shown in Table \ref{table:datasets_stats}.
\begin{table}[t]
  \caption{Partitioning of dataset into train and test. For each categorization approaches, we show the distribution of the four CS classes \{C0,C1,C2,C3\} for train and test sets.}
  \label{table:datasets_stats}
  \centering
  \setlength\tabcolsep{2pt}
\begin{tabular}{|l|r|r|r|r|r|r|r|r|r|r|r|r|}
\cline{2-13}
\multicolumn{1}{c|}{}&\multicolumn{4}{c|}{} & \multicolumn{4}{c|}{\textbf{K-means}} & \multicolumn{4}{c|}{\textbf{K-means}}
\\
\multicolumn{1}{c|}{}&\multicolumn{4}{c|}{\textbf{CMI}} & \multicolumn{4}{c|}{\textbf{[CMI]}} & \multicolumn{4}{c|}{\textbf{[CMI+CS\%+En\%]}}
\\\cline{2-13}
\multicolumn{1}{c|}{} & \multicolumn{1}{c|}{\textbf{C0}} & \multicolumn{1}{c|}{\textbf{C1}} & \multicolumn{1}{c|}{\textbf{C2}} & \multicolumn{1}{c|}{\textbf{C3}} & \multicolumn{1}{c|}{\textbf{C0}} & \multicolumn{1}{c|}{\textbf{C1}} & \multicolumn{1}{c|}{\textbf{C2}} & \multicolumn{1}{c|}{\textbf{C3}} & \multicolumn{1}{c|}{\textbf{C0}} & \multicolumn{1}{c|}{\textbf{C1}} & \multicolumn{1}{c|}{\textbf{C2}} & \multicolumn{1}{c|}{\textbf{C3}}\\\hline
\textbf{Train} & 28.9\% & 46.7\% & 15.6\% & 8.9\% & 73.3\% & 17.8\% & 6.7\% & 2.2\% & 62.2\% & 22.2\% & 13.3\% & 2.2\%\\\hline
\textbf{Test} & 40.0\% & 25.0\% & 20.0\% & 15.0\% & 65.0\% & 20.0\% & 5.0\% & 10.0\% & 60.0\% & 25.0\% & 5.0\% & 10.0\%\\\hline
\textbf{Overall} & 32.3\% & 40.0\% & 16.9\% & 10.8\% & 70.8\% & 18.5\% & 6.2\% & 4.6\% & 61.5\% & 23.1\% & 10.8\% & 4.6\% \\\hline
\end{tabular}
\end{table}

\section{Experimental Results}
In this Section, we evaluate the correlations between the sociological and psychological factors and CS behaviour. We first investigate significant correlations using Pearson Correlation. Afterwards, we report the results of our machine learning models and look into the importance of the factors in the modeling process.
\subsection{Pearson Correlation}
\begin{figure}[h]
  \centering
  \caption{Heatmap of features and CS metrics.
  }
  \label{fig:heatmap}
  \resizebox{0.6\linewidth}{!}{
  \includegraphics[width=\linewidth]{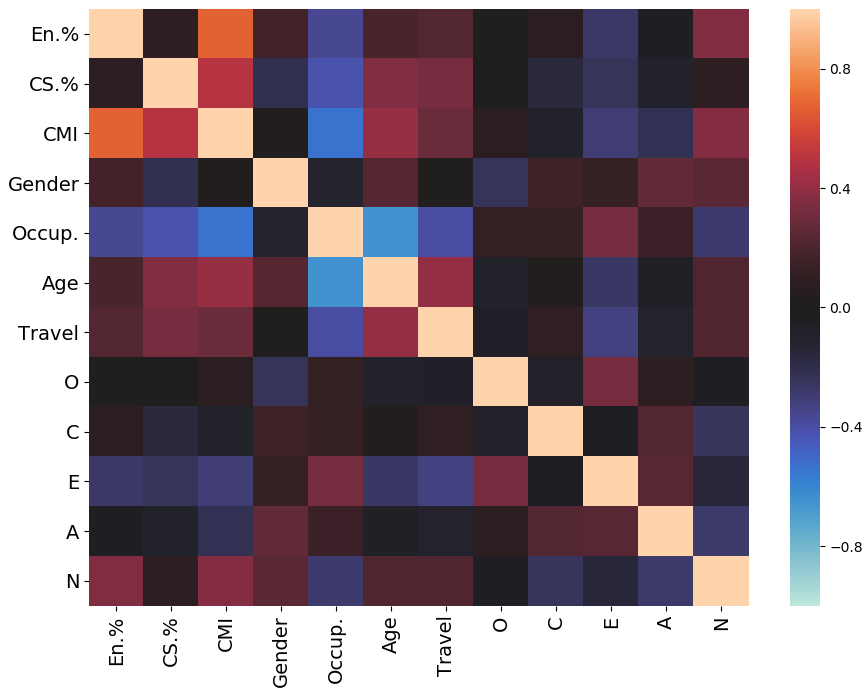}
  }
\end{figure}
Correlations between CMI and the factors under investigation were analyzed using Pearson Correlation Coefficients with a significance criterion of $p < .05$* and $p < .01$**, two-tailed tests. The significant correlations were identified as: occupation** (-0.53), age** (0.41), neuroticism** (0.36), traveling experiences* (-0.30), and extraversion* (-0.30). Figure \ref{fig:heatmap} presents the correlations between the factors and the three CS metrics using heatmaps. We order the factors with their influence degree on CS behaviour as follows: occupation, age, neuroticism, traveling experiences, extraversion, gender, agreeableness, conscientiousness, and openness.

\subsection{Predictive Models}
\begin{table}[t]
  \caption{The accuracy of the predictive models on test set for the classification task.}
  \label{table:ML_results_classification}
  \centering
  \setlength\tabcolsep{1pt}
\begin{tabular}{|l|l|l|l|}
\cline{2-4}
 \multicolumn{1}{c|}{}& \multicolumn{3}{c|}{\textbf{Categorization Approach}} \\ \hline
\multicolumn{1}{|c|}{} &  & \multicolumn{1}{|c|}{\textbf{K-means}} & \multicolumn{1}{|c|}{\textbf{K-means}} \\
\multicolumn{1}{|c|}{\textbf{Algorithm}} & \multicolumn{1}{|c|}{\textbf{CMI}} & \multicolumn{1}{|c|}{\textbf{[CMI]}} & \multicolumn{1}{|c|}{\textbf{[CMI+CS\%+En\%]}} \\\hline
\textbf{Random Trees} & \textbf{0.55} & 0.55 & 0.60 \\ \hline
\textbf{Random Forest} & 0.20 & 0.45 & 0.50 \\ \hline
\textbf{XGBoost Tree} & 0.40 & 0.60 & \textbf{0.65} \\ \hline
\textbf{XGBoost Linear} & 0.40 & \textbf{0.75} & 0.60 \\ \hline
\textbf{LSVM} & 0.40 & 0.65 & 0.60 \\ \hline
\end{tabular}
\end{table}

\begin{figure*}[t]%
\centering
\caption{Confusion Matrices for the best-performing model for each categorization approach on test sets for the classification task.}
\label{fig:confusion_matrix}
   \resizebox{\linewidth}{!}{
\begin{subfigure}{.55\columnwidth}
\includegraphics[width=\columnwidth]{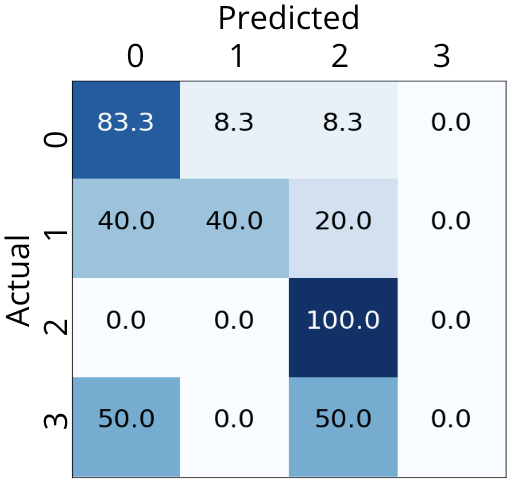}%
\caption{CMI}%
\label{subfiga}%
\end{subfigure}\hfill%
\begin{subfigure}{.5\columnwidth}
\includegraphics[width=\columnwidth]{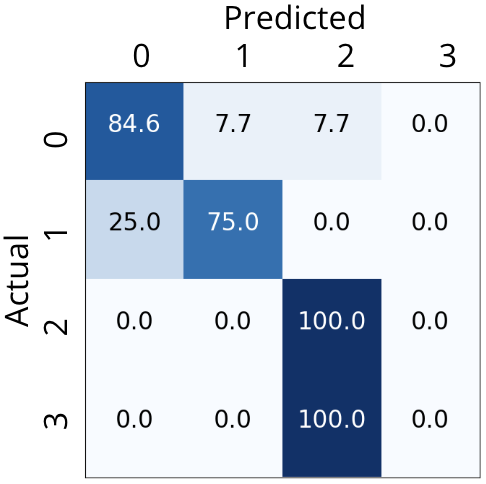}%
\caption{K-means[CMI]}%
\label{subfigb}%
\end{subfigure}\hfill%
\begin{subfigure}{.5\columnwidth}
\includegraphics[width=\columnwidth]{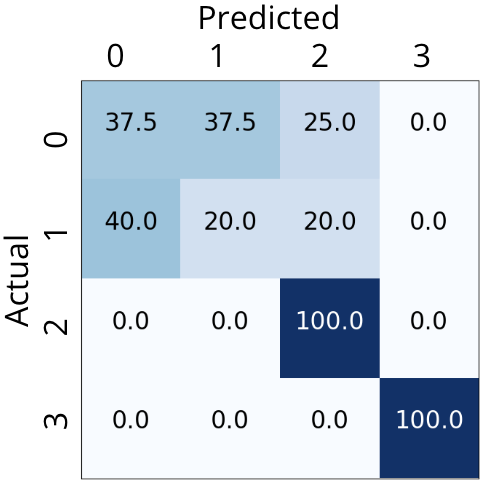}%
\caption{K-means[CMI+CS\%+En\%]}%
\label{subfigc}%
\end{subfigure}%
}
\end{figure*}
We verify the correlations by training ML models to predict CS behaviour given the participants' meta-data. We analyze the output of the ML models in terms of the models' ability to predict CS levels. We also show the predictor importance chart, indicating the relative importance of each predictor in the modeling process, 
thus identifying the influential factors for predicting CS levels.
\subsection{Classification Task}
\begin{figure}[h]
  \caption{The predictor importance chart for the classification task.}
  \label{fig:predictor_imp_classification}
  \includegraphics[width=0.8\linewidth]{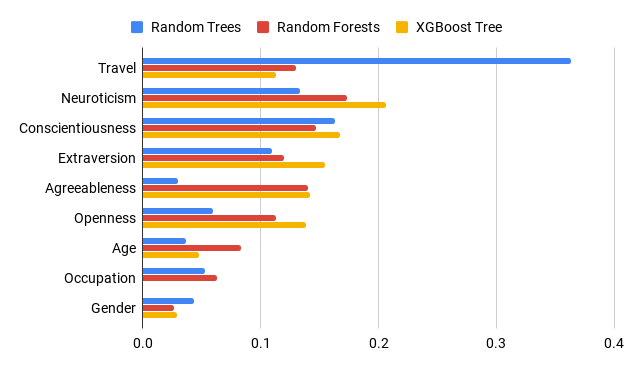}
\end{figure}

We report the classification accuracy of the models for each categorization approach in Table \ref{table:ML_results_classification}. We show that the ML models are able to correctly predict CS classes with an accuracy higher than 55\%. In Figure \ref{fig:confusion_matrix}, we present the confusion matrices for the best-performing models for each categorization approach on test sets. The rows represent the actual classes as specified by the categorization approach, and the columns represent the classes generated by our predictive models. By looking into the confusion matrices, we find that 75\% of the incorrectly classified samples were assigned to neighbouring classes. 
The predictor importance chart for Random Trees, Random Forest, and XGBoost Tree is shown in Figure \ref{fig:predictor_imp_classification}. It can be seen that the highest influential factors are the traveling experiences and personality traits. 

\subsection{Regression Task}
We also show results for the regression task, where we train ML models to predict users' CMI values as well as percentage of English embedded words based on their profiles. The experiments of the regression task are mainly conducted for the purpose of looking into the predictor importance. It is to be noted that the aim of these experiments is not to achieve high-performing ML models that can predict CMI values, as (1) predicting CS values is a challenging task and while predicting a user class can be a reasonable task, predicting an exact value is a hard task, even for humans and (2) for the purpose of building user-adaptive NLP models, the models would be adapted for each class of users, thus the exact value would not be required for this purpose.\\

\begin{figure}[th!]
\centering
\caption{The predictor importance charts for the regression tasks.}
\label{fig:predictor_imp_regression}
\begin{subfigure}{0.8\textwidth}
    \includegraphics[width=\textwidth]{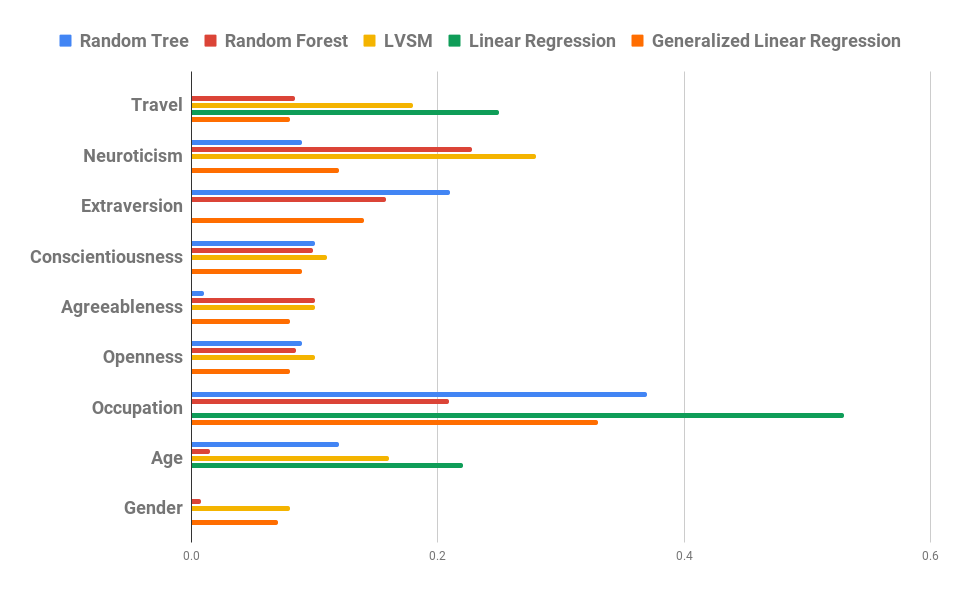}
    \caption{Predicting CMI}
    \label{fig:first}
\end{subfigure}
\hfill
\begin{subfigure}{0.8\textwidth}
    \includegraphics[width=\textwidth]{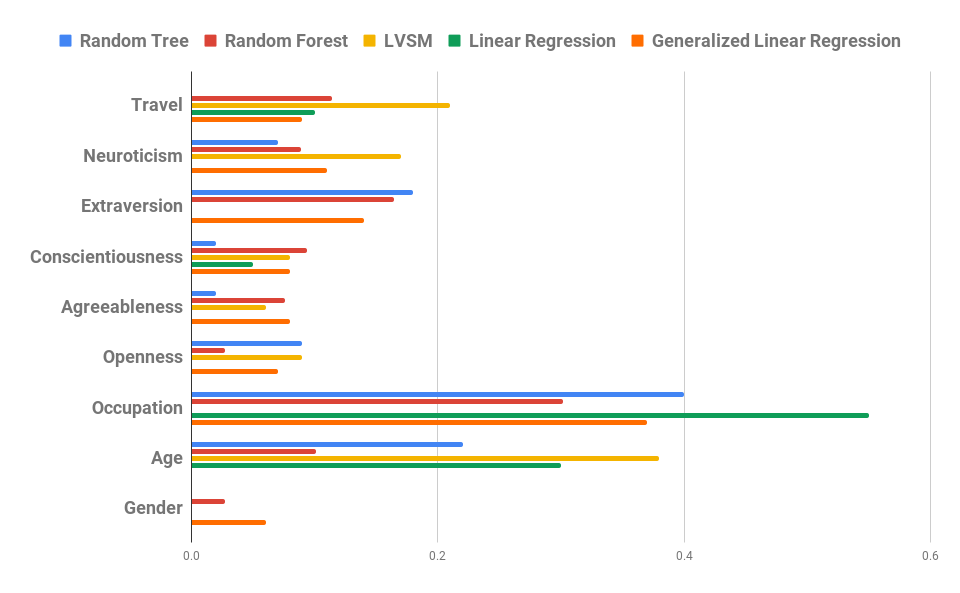}
    \caption{Predicting English Percentage}
    \label{fig:second}
\end{subfigure}
\label{fig:figures}
\end{figure}

For the task of predicting CMI, the models achieve an absolute mean error of 0.082-0.089 as shown in Table \ref{table:ML_results_regression_CMI}. For the task of predicting English percentage, the models achieve an absolute mean error of 8.2-9.5\%, as shown in Table \ref{table:ML_results_regression_En}. In Figure \ref{fig:predictor_imp_regression}, We show the predictor importance charts for both tasks across the different ML models. By averaging the predictor importance values across the models, we identify the top influential factors for the task of predicting CMI as: occupation, neuroticism, travel, age, and extraversion, and for the task of predicting English percentage as: occupation, age, travel, extraversion, and neuroticism.
\begin{table}[t]
  \caption{The accuracy of the predictive models on test set for the regression task of predicting CMI.}
  \label{table:ML_results_regression_CMI}
  \centering
  \setlength\tabcolsep{2pt}
\begin{tabular}{|l|r|r|r|r|r|}
\hline
& \multicolumn{1}{c|}{\textbf{Min.}} & \multicolumn{1}{c|}{\textbf{Max.}} & \multicolumn{1}{c|}{\textbf{Mean}} & \multicolumn{1}{c|}{\textbf{Mean Absolute}} &  \\
\multicolumn{1}{|c|}{\textbf{Algorithm}}& \multicolumn{1}{c|}{\textbf{Error}} & \multicolumn{1}{c|}{\textbf{Error}} & \multicolumn{1}{c|}{\textbf{Error}} & \multicolumn{1}{c|}{\textbf{Error}} &\multicolumn{1}{c|}{\textbf{STDEV}} \\\hline
 \textbf{Random Tree} & -0.096& 0.285 & 0.024 &0.089&0.116\\ \hline
\textbf{Forest Forest} & -0.127 & 0.244 & 0.006 & \textbf{0.082}&0.107\\ \hline
\textbf{LSVM} & -0.144 & 0.219 & 0.013 & 0.087&0.105\\ \hline
\textbf{Linear Regression} & -0.159 & 0.227 & 0.008 & \textbf{0.082}&0.108\\ \hline
\textbf{Generalized Linear Regression} & -0.164 & 0.234 & 0.015 & 0.089&0.112\\ \hline
\end{tabular}
\end{table}

\begin{table}[t]
  \caption{The accuracy of the predictive models on test set for the regression task of predicting percentage of English words.}
  \label{table:ML_results_regression_En}
  \centering
  \setlength\tabcolsep{2pt}
\begin{tabular}{|l|r|r|r|r|r|}
\hline
& \multicolumn{1}{c|}{\textbf{Min.}} & \multicolumn{1}{c|}{\textbf{Max.}} & \multicolumn{1}{c|}{\textbf{Mean}} & \multicolumn{1}{c|}{\textbf{Mean Absolute}} & \\
\multicolumn{1}{|c|}{\textbf{Algorithm}}& \multicolumn{1}{c|}{\textbf{Error}} & \multicolumn{1}{c|}{\textbf{Error}} & \multicolumn{1}{c|}{\textbf{Error}} & \multicolumn{1}{c|}{\textbf{Error}} & \multicolumn{1}{c|}{\textbf{STDEV}} \\\hline
 \textbf{Random Tree} & -10.67 & 29.90 & 2.31 & 9.18 & 11.91\\ \hline
\textbf{Forest Forest} & -17.50 & 23.85 & 1.12 & 9.47 & 11.86\\ \hline
\textbf{LSVM} & -17.25 & 22.70 & 2.17 & 9.40 & 11.10\\ \hline
\textbf{Linear Regression} & -19.53 & 22.29 & 0.13 & \textbf{8.20} & 10.89\\ \hline
\textbf{Generalized Linear Regression} & -17.22 & 23.24 & 1.29 & 9.17 & 11.46\\ \hline
\end{tabular}
\end{table}

\section{Discussion}
We have investigated the influence of several factors on CS behaviour using Person Correlation Coefficients and predictor importance in the ML modeling process. Although both approaches did not show the same importance across all factors, there are common findings. In opposition to previous work \cite{BBS15}, gender is not identified as an influential factor. The following personality traits also did not show any correlation with CS: Conscientiousness, Agreeableness, and Openness. Influential factors are identified to be traveling experiences, Neuroticism and Extraversion, occupation, and age.\\

While \cite{DW14} report that Neuroticism is negatively linked to CS levels, our findings, on the contrary, show a positive correlation. We believe both findings are plausible. On one hand, as suggested by \cite{Mol07}, individuals with high Neuroticism are likely to experience more anxiety in the CS process than individuals with low Neuroticism. On the other hand, individuals with high Neuroticism are more likely to lack confidence, and possibly resort to CS to boost their confidence, where CS is attributed to the following scenarios: reflecting a certain socioeconomic identity which can give the speaker more credibility and reliability \cite{Ner11}, persuading an audience \cite{Hol17}, and reflecting social status \cite{Eld14}. Regarding Extraversion, \cite{DL14} report a positive link between Extraversion and self-reported CS. 
In the scope of our study, we confirm that Extraversion influences CS levels, however, according to the results using Pearson Correlation, we report a negative correlation. 
It is first to be noted that in both studies \cite{DW14,DL14}, the authors look into the correlation between personality traits and self-reported CS levels, while we use actual CS levels. It is also to be noted that the experiments were conducted in different countries. 
Given that the reason of CS evolution varies across countries and cultures, the factors affecting CS might not necessarily hold. For examples, CS in Egypt evolved due to the middle and upper classes being enrolled in private and international schools and universities, and thus English proficiency became an indication of social and educational class, thus CS can be used to reflect certain socio-economic status. This might not be the case for other countries, where CS could have evolved due to other reasons, such as immigration.\\

While occupation and age are reasonable factors to affect CS, under this setting, we believe that the effect caused by age and occupation on the CS level actually falls under Participant Roles and Relationship external factor, where the participants in the older age group are university employees and the younger age group are students. Given that both interviewers were undergraduate students, the CS level of participants could have been affected by the interviewer-interviewee relationship, as university employees could be used to using English when communicating with students, given that the experiment was held in a private university, where English is the official instruction language.\\

We also discuss two main factors that limit the accuracy of our models. The first factor is the limited number of participants. In order to get accurate results, we relied on actual CS metrics obtained from speech transcriptions rather than self-reported levels. Given that corpus collection is time- and money-consuming, we could not collect more data. Secondly, in order to fully model a user's CS behaviour, all factors affecting CS need to be taken into account, requiring a large-scale experiment covering diverse situations, sociological and psychological profiles. Therefore, modeling CS behaviour is a hard task. Despite the small sample size, both qualitative and experimental results show the strong correlation between character traits and CS levels, highlighting the potential benefits of incorporating character traits in user-adapted CS NLP applications. In this paper, we focus on a few factors affecting CS behaviour, with a specific interest on the personality side, as this area has received considerably less attention than other factors. For future work, it would be interesting to investigate the effect of personality traits across different socio-economic standards. Further investigations are also needed regarding the effect of travel experiences, where more details are provided about the language that was used by participants during their stay.
\section{Conclusion}
Predicting users' CS levels can allow for user-adapted modeling in NLP tasks, paving the way for developing more accurate NLP systems. 
In this work, we investigate predicting users' CS levels based on their character profiles. We conduct an empirical study where we interview Arabic-English bilinguals and collect their CS levels and profiles. We use several ML algorithms, to build predictive models. The results show that ML algorithms can be leveraged to learn users' CS level from their profile. Finally, we identify the factors that dominate the prediction process, and thus contribute the most to CS behaviour. We report that CS is found to be affected by participant roles and relationship external factors, which is reflected in correlations with age and occupation. We also report traveling experiences in addition to Neuroticism and Extraversion personality traits among the top influential factors. 
\appendix
\section{Questionnaire}
\label{sec:appendix-questionnaire}
\begin{enumerate}
    \item Gender
        \begin{itemize}
            \item Female
            \item Male
        \end{itemize}
    \item Age group
        \begin{itemize}
            \item 18-23
            \item 23-35
            \item Above 35
        \end{itemize}
    \item Which language do you consider your mother-tongue?
        \begin{itemize}
            \item Pure Arabic
            \item Pure English
            \item Code-switched, Arabic-English
        \end{itemize}
    \item What was your longest stay in a foreign country for tourism/ education?
        \begin{itemize}
            \item I didn’t travel before.
            \item less than a month.
            \item 1-3 months.
            \item 3-6 months.
            \item 6-12 months.
            \item 1-3 years.
            \item more than 3 years.
        \end{itemize}
    \item Does your family speak multiple languages?
        \begin{itemize}
            \item Yes
            \item No
        \end{itemize}
    \item Do your close friends speak multiple languages?
        \begin{itemize}
            \item Yes
            \item No
        \end{itemize}
    \item Were you in a national or an international school?        \begin{itemize}
            \item National
            \item International
        \end{itemize}
    \item Did you have classes in school where you had to speak the English language?
        \begin{itemize}
            \item Yes
            \item No
            \item Maybe
        \end{itemize}
    \item On scale 1 to 5, Are you aware that you code-switch (Switch between multiple languages) within your normal speech?
    \item On scale 1 to 5, how often do you code-switch?
    \item On scale 1 to 5, how often do your friends code-switch?
    \item On scale 1 to 5, code-switching is done due to strength in both languages?
    \item On scale 1 to 5, code-switching is done due to weakness in the common spoken language?
\end{enumerate}

\section*{Acknowledgment}
We would like to thank Mohamed Elmahdy for his valuable suggestions as well as Dahlia Sabet, Amira Khaled, and Karim Elkhafif for helping us in collecting the corpus. Special thanks also goes to all the participants who volunteered to help us with our project.
\bibliographystyle{splncs04}
\bibliography{main}
\end{document}